\documentclass{article}

\usepackage[preprint]{neurips_2023}

\usepackage[utf8]{inputenc} 
\usepackage[T1]{fontenc}    
\usepackage{hyperref}       
\usepackage{url}            
\usepackage{booktabs}       
\usepackage{amsfonts}       
\usepackage{nicefrac}       
\usepackage{microtype}      
\usepackage{xcolor}         
\usepackage{multirow}
\usepackage{graphicx}
\usepackage{subcaption}

\usepackage{amsmath,amsfonts,bm}

\def\eqref#1{equation~\ref{#1}}

\def\1{\bm{1}}

\def\vs{{\bm{s}}}

\def\mA{{\bm{A}}}

\def\mP{{\bm{P}}}
\def\mQ{{\bm{Q}}}

\def\mT{{\bm{T}}}

\DeclareMathAlphabet{\mathsfit}{\encodingdefault}{\sfdefault}{m}{sl}
\SetMathAlphabet{\mathsfit}{bold}{\encodingdefault}{\sfdefault}{bx}{n}

\DeclareMathOperator{\integer}{int}

\title{Mixture of Quantized Experts (MoQE):\\ Complementary Effect of \\ Low-bit Quantization and Robustness}

\author{%
  Young Jin Kim\thanks{Equal contribution.} \\
  Microsoft \\
  \texttt{youki@microsoft.com} \\
  \And
  Raffy Fahim\footnotemark[1] \\
  Microsoft \\
  \texttt{raffybekheit@microsoft.com} \\
  \AND
  Hany Hassan Awadalla \\
  Microsoft \\
  \texttt{hanyh@microsoft.com} \\
}

\begin{document}

\maketitle

\begin{abstract}
Large Mixture of Experts (MoE) models could achieve state-of-the-art quality on various language tasks, including machine translation task, thanks to the efficient model scaling capability with \textit{expert parallelism} \citep{fedus2021switch}. However, it has brought a fundamental issue of larger memory consumption and increased memory bandwidth bottleneck at deployment time. In this paper, we propose \textit{Mixture of Quantized Experts (MoQE)} which is a simple \textit{weight-only} quantization method applying ultra low-bit down to 2-bit quantizations only to expert weights for mitigating the increased memory and latency issues of MoE models. We show that low-bit quantization together with the MoE architecture delivers a reliable model performance while reducing the memory size significantly even without any additional training in most cases. In particular, \texttt{expert} layers in MoE models are much more robust to the quantization than conventional feedforward networks (FFN) layers. In our comprehensive analysis, we show that MoE models with 2-bit expert weights can deliver better model performance than the dense model trained on the same dataset. As a result of low-bit quantization, we show the model size can be reduced by 79.6\% of the original half precision floating point (fp16) MoE model. Combined with an optimized GPU runtime implementation, it also achieves 1.24X speed-up on A100 GPUs.
\end{abstract}

\section{Introduction}
The Mixture-of-Experts (MoE) architecture efficiently increase the number of model parameters, while maintaining a sub-linear increase in computational requirements by activating only a few small number of experts at a time \citep{lepikhin2020gshard, fedus2021switch, kim2021scalable, artetxe2021efficient}. As a result, MoE models could achieve higher quality compared to the dense models by increasing the size of the model dramatically. In a large scale distributed training setting, this can be efficiently scaled with expert parallelism\citep{fedus2021switch}. However, during inference scenarios, despite the sub-linear increase in computational load, there is a notable surge in memory bandwidth requirement. Table \ref{inference-speed-baseline} shows that how much memory bandwidth overhead is introduced, even when employing just 32 experts without a corresponding increase in theoretical FLOPs, as implemented with top-1 gating \citep{fedus2021switch} on an NVIDIA A100 GPU.

\begin{table}[ht]
\caption{Inference speed measurements and model sizes of dense and MoE models. Both models run with batch size of 24 and the throughput is measured with the number of sentences processed for one second.}
\label{inference-speed-baseline}
\begin{center}
\begin{tabular}{l|r|r|r}
\hline 
\multirow{2}{*}{Model}  &{Throughput} & Model size & \% of MoE weights \\ 
&{(sentences/second)} & (\textit{fp16}) in GB \\ 
\hline
Dense         & 14.02 & 1.18 & - \\
MoE (32 experts)             & 5.37 & 9.91 & \textbf{92.8 \%}\\
\hline
Difference             & 0.38X & 8.38X &  - \\
\hline
\end{tabular}
\end{center}
\end{table}

In spite of the progress on the training of MoE models, there have been only a few handfuls of studies related to MoE model inference. \citet{Rajbhandari2022DeepSpeedMoEAM} designs a more efficient MoE architecture and distributed runtime. \citet{Kudugunta2021BeyondDT} uses task specific information to reduce the size of the model at deployment time by only loading task specific experts. \citet{kim2021scalable} prunes some experts at deployment time to reduce the model size by trading-off model performance. \citet{zoph2022designing} uses knowledge distillation technique to distill a large MoE model into a smaller dense model to reduce the memory consumption and improve the throughput. Even with all the proposed techniques, there has not been a solution to accelerate the inference of MoE models while maintaining the accuracy.

To effectively solve the problem, we empirically show that expert weights are highly robust to the quantization, therefore they can be quantized to 3-bit without additional training or calibration data and to 2-bit with Quantization Aware Training (QAT) which results in 79.6\% reduction in memory size. Also with a runtime optimization, we show that the method boosts the inference speed more than 1.24X faster on A100 GPUs.

\section{Quantization robustness of MoE layers}
\label{sec:quant_methods}

\subsection{Numerical distribution of model weights}
While quantizing matrices, outliers usually skew the range to be quantized and scaling factors get too large and result in poor quantization quality. We investigate if outliers exist in MoE and other layers.

\begin{figure}[h]
    \centering
    \begin{subfigure}[b]{0.49\textwidth}
        \centering
        \includegraphics[width=1.0\textwidth]{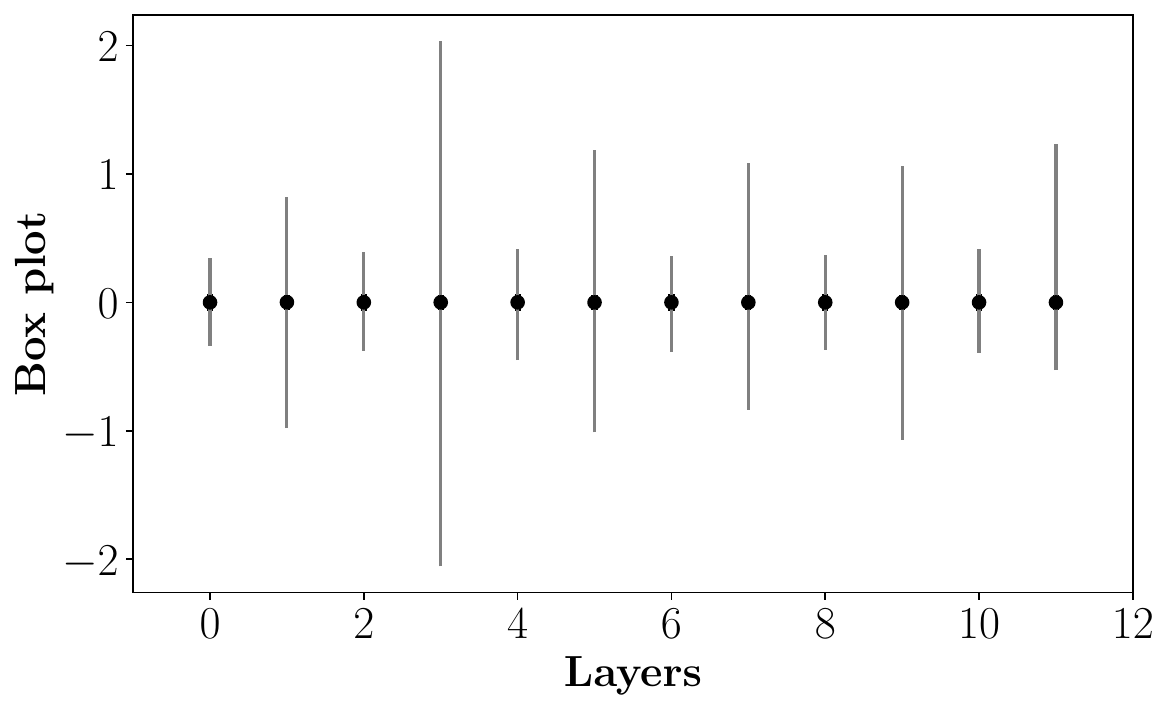}
        \vskip -0.1in
        \caption{\centering FFN linear 1 weight distribution across layers}
        \label{fig:fc1-layer-stats}
        \end{subfigure}
    \begin{subfigure}[b]{0.49\textwidth}
        \centering
        \includegraphics[width=1.0\textwidth]{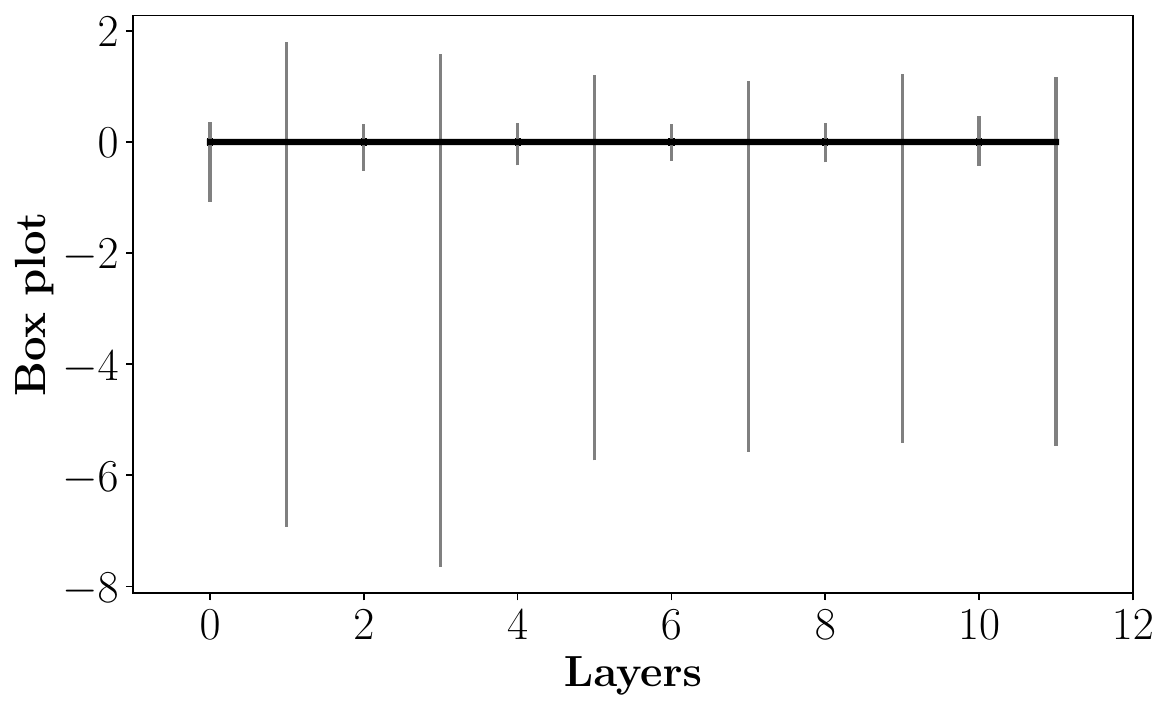}
        \vskip -0.1in
        \caption{\centering FFN linear 2 weight distribution across layers}
        \label{fig:fc2-layer-stats}
    \end{subfigure}
    \caption{FFN weight distribution across layers. Even number layers $\{0, 2, ...\}$ are expert FFN layers and odd number layers $\{1, 3, ...\}$ are normal dense FFN layers. (a) shows the first linear layer in FFN and (b) shows the second linear layer in FFN.}
    \label{fig:weight-distribution-layer}
\end{figure}

Figure \ref{fig:weight-distribution-layer} shows weight distribution box plots of linear layers in the MoE model's FFN blocks. We use a normal two layer FFN block from the Transformer paper \citep{Vaswani2017AttentionIA}. Following the widely used practice, an MoE layer is in every other layer \citep{lepikhin2020gshard,fedus2021switch,kim2021scalable}. Even number layers $\{0, 2, ...\}$ are expert FFN layers and odd number layers $\{1, 3, ...\}$ are normal dense FFN layers. From the plot, dense FFN layers have a much larger range than MoE FFN layers. This indicates that dense FFN layers have more outliers than MoE FFN layers. This phenomenon is more prevalent in the second linear layers sometimes reaching down to $-8.0$ which is shown in Figure \ref{fig:fc2-layer-stats}.



\subsubsection{Robustness of expert layers to quantization}
\label{sec:sensitivity}
To better understand how applying quantization on different parts of an MoE model affects the accuracy, we conduct a set of experiments with various quantization bits. We divide an MoE model into four parts: (i) expert FFNs, (ii) dense FFN layers, (iii) self-attention layers and (iv) cross-attention layers. Based on the observation that linear quantization works better with lower bits, we use it for this set of experiments. 

\begin{figure}[h]
    \centering
    \includegraphics[width=0.7\textwidth]{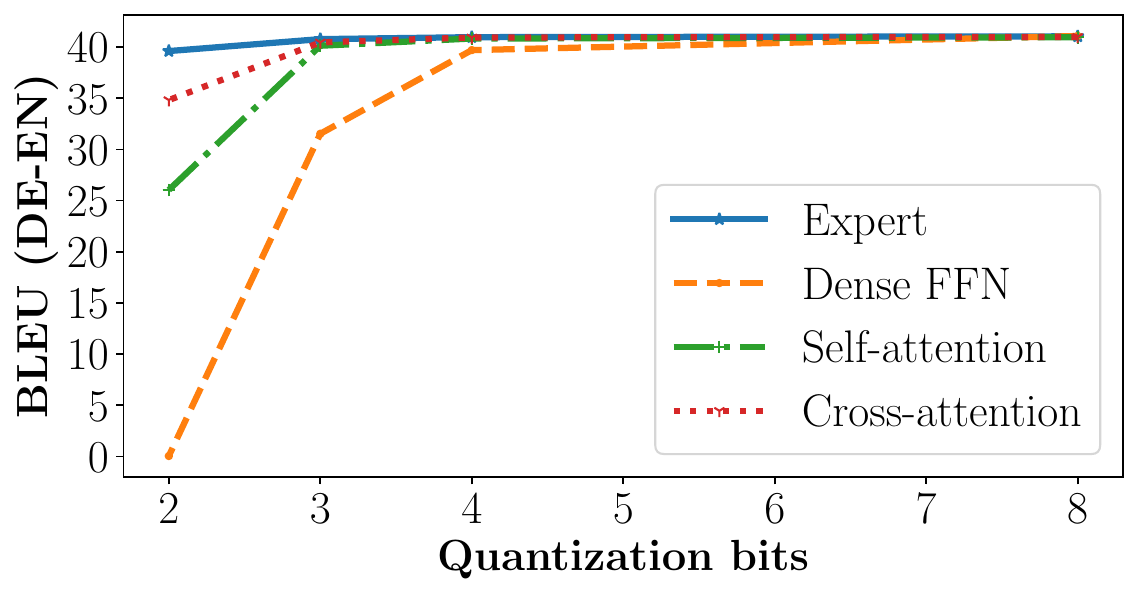}
    \vskip -0.1in
    \caption{Quantization impact on different MoE model parts (channel-wise linear quantiztation without any additional training).}
    \label{fig:different NN parts quant}
\end{figure}

Figure \ref{fig:different NN parts quant} shows evaluation BLEU~\footnote{https://github.com/mjpost/sacrebleu} scores which is one of the quality metrics for machine translation when quantizing different parts of the MoE model. We observe that quantizing expert FFN layers to 2-bit does not seriously impact the overall model quality. However, quantizing other parts of the model into 2-bit hurts the output quality significantly. Quantized cross-attention and self-attention blocks still can maintain the quality with 3-bit quantization, but their performance gets impacted with 2-bit quantization. On the other hand, dense FFN layers get significant impact with lower bit quantization of 2-bit and 3-bit. With 3-bit quantization, the model score drops 23 \% of original score, and 2-bit quantization on dense FFN layers gives almost zero score. We also include the same study on a dense model in Appendix \ref{app:dense-layers}, the similar pattern with 2 and 3 bit quantization is observed.

\section{Experiments}
Given the observations from the previous section, we aggressively apply low-bit quantization on MoE weights only which result in MoQE (Mixture-of-Quantized-Experts). We use multilingual machine translation task for our experiments. The details of the datasets, quality metrics and model architectures are described in Appendix \ref{app:setup}

\subsection{MoQE performance results}
We apply MoQE quantization recipe to an MoE model and compare the performance with the baseline MoE model in Table \ref{tab:mose-result}. For a reference, a dense model is also trained on the same dataset as the MoE model. For the MoE model, various quantization settings ranging from 8-bit to 2-bit are measured together with the original fp16 performance. For 2-bit quantization, additional QAT is applied.

First of all, the MoE model achieves 2.87\% improvement on the BLEU score while increasing the model size to 8.38X of the original dense model. When 4-bit post-training quantization is applied, it still maintains 2.11\% higher BLEU score than the original dense model. This reduces the memory consumption by 68\% and while speeding up inference 1.24X faster than fp16 MoE model. With 2-bit QAT, the MoE model can still maintain 1.88\% higher quality than the original dense model, but the model size is now only 1.71X of the original dense model. 


\begin{table}[t]
\caption{The model performance comparison. All the models are trained on same data up to the convergence with 200,000 update steps. The baseline is the FLOPs equivalent dense model's BLEU score and speed.}
\label{tab:mose-result}
\begin{center}
\begin{tabular}{c|ccccccccccc}
\hline
\multirow{2}{*}{\bf Model type} & \multirow{2}{*}{\bf Precision} & {\bf Average BLEU} & {\bf Throughput}   & {\bf Size}   \\
& &  (difference \%) & (X times) & (X times)  \\
\hline
Dense & fp16 & 45.06 (0) & - & - \\
\hline
 MoE Baseline & fp16  & {\bf 46.35 (+2.87)} & 1.00X & 1.00X  \\
\hline
\multirow{3}{*}{MoE 5.3B (32 experts)} & int8  & {\bf 46.34 (+2.85)} & 1.16X & 0.54X  \\\cline{2-6}
& int4  & 46.18 (+2.49) & {\bf 1.24X} & 0.32X  \\\cline{2-6}
MoQE & int3  & 46.01 (+2.11) & Not optimized & 0.26X  \\\cline{2-6}
& int2  & \multirow{2}{*}{45.90 (+1.88)} & \multirow{2}{*}{Not optimized} & \multirow{2}{*}{0.20X}  \\
& QAT  & &  &  &  \\
\hline
\hline
\end{tabular}
\end{center}
\end{table}

\subsection{Robustness comparison between MoE and dense models}
We compare robustness against low-bit quantization between MoE and dense models using the post-training quantization without any QAT. For the dense model, quantization with different bits is applied to the even numbered FFN layers. Appendix \ref{app:dense-layers} shows this is the best layer selection for the dense model. We use two different datasets to verify the proposed quantization method works in different model settings.
\begin{figure}[h]
    \centering
    \includegraphics[width=0.6\textwidth]{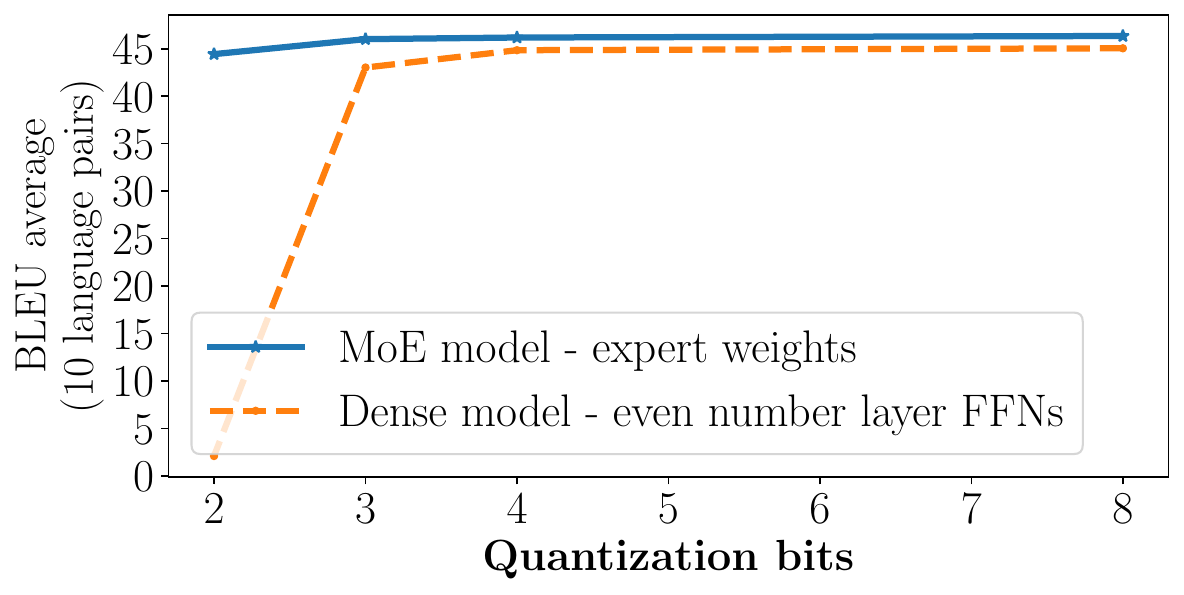}
    \vskip -0.1in
    \caption{Quantization performance comparison between MoE and dense models. 10 different language pair scores are averaged.}
    \label{fig:dense-moe-euro}
\end{figure}

Figure \ref{fig:dense-moe-euro} presents the experiment with the model trained with 20 direction multilingual translation dataset. It shows the average BLEU scores with different quantization precision for both MoE and dense models. The MoE model can maintain accuracy within -0.3 down to 3-bit and -1.82 for 2-bit. On the other hand, the dense model can preserve the accuracy only down to 4-bit, but starts to lose significant accuracy more than 2 BLEU scores when it goes down to 3-bits. In case of 2-bits, dense model loses most of capability by -42.96 BLEU scores. Table \ref{tab:moe-dense} in Appendix shows the score differences by quantization for both MoE and dense models on 10 different language pairs translations.

\section{Conclusions and limitations}
This paper shows how much MoE models are robust to the low-bit quantization with various experiments. By analyzing component-wise sensitivity and various quantization design choices, we present an efficient and effective way to reduce the model size which results in 4.9X model size reduction. With an optimized runtime, 4-bit quantized model can run 1.24X faster than the fp16 model.

Even with the interesting findings, the study has a few limitations. First of all, there does not exist an optimized implementation for lower than 4-bit quantization, yet. This is a good potential future research direction. Secondly, 2-bit quantization still requires QAT while 3-bit or higher bit quantization does not. Lastly, there could be a hybrid approach to mix different quantization precisions between MoE layers and the other layers which can result in more optimal model performance.

\clearpage




\bibliography{iclr2023_conference}

\begin{thebibliography}{15}
\providecommand{\natexlab}[1]{#1}
\providecommand{\url}[1]{\texttt{#1}}
\expandafter\ifx\csname urlstyle\endcsname\relax
  \providecommand{\doi}[1]{doi: #1}\else
  \providecommand{\doi}{doi: \begingroup \urlstyle{rm}\Url}\fi

\bibitem[Aji \& Heafield(2020)Aji and Heafield]{Aji2020CompressingNM}
Alham~Fikri Aji and Kenneth Heafield.
\newblock Compressing neural machine translation models with 4-bit precision.
\newblock In \emph{NGT}, 2020.

\bibitem[Artetxe et~al.(2021)Artetxe, Bhosale, Goyal, Mihaylov, Ott, Shleifer, Lin, Du, Iyer, Pasunuru, et~al.]{artetxe2021efficient}
Mikel Artetxe, Shruti Bhosale, Naman Goyal, Todor Mihaylov, Myle Ott, Sam Shleifer, Xi~Victoria Lin, Jingfei Du, Srinivasan Iyer, Ramakanth Pasunuru, et~al.
\newblock Efficient large scale language modeling with mixtures of experts.
\newblock \emph{arXiv preprint arXiv:2112.10684}, 2021.

\bibitem[Fedus et~al.(2021)Fedus, Zoph, and Shazeer]{fedus2021switch}
William Fedus, Barret Zoph, and Noam Shazeer.
\newblock Switch transformers: Scaling to trillion parameter models with simple and efficient sparsity.
\newblock \emph{arXiv preprint arXiv:2101.03961}, 2021.

\bibitem[Kasai et~al.(2021)Kasai, Pappas, Peng, Cross, and Smith]{Kasai2021DeepES}
Jungo Kasai, Nikolaos Pappas, Hao Peng, James Cross, and Noah~A. Smith.
\newblock Deep encoder, shallow decoder: Reevaluating non-autoregressive machine translation.
\newblock In \emph{ICLR}, 2021.

\bibitem[Ke et~al.(2021)Ke, He, and Liu]{Ke2021RethinkingPE}
Guolin Ke, Di~He, and Tie-Yan Liu.
\newblock Rethinking positional encoding in language pre-training.
\newblock \emph{ArXiv}, abs/2006.15595, 2021.

\bibitem[Kim et~al.(2019)Kim, Junczys-Dowmunt, Hassan, Aji, Heafield, Grundkiewicz, and Bogoychev]{kim2019research}
Young~Jin Kim, Marcin Junczys-Dowmunt, Hany Hassan, Alham~Fikri Aji, Kenneth Heafield, Roman Grundkiewicz, and Nikolay Bogoychev.
\newblock From research to production and back: Ludicrously fast neural machine translation.
\newblock In \emph{Proceedings of the 3rd Workshop on Neural Generation and Translation}, pp.\  280--288, 2019.

\bibitem[Kim et~al.(2021)Kim, Awan, Muzio, Salinas, Lu, Hendy, Rajbhandari, He, and Awadalla]{kim2021scalable}
Young~Jin Kim, Ammar~Ahmad Awan, Alexandre Muzio, Andres Felipe~Cruz Salinas, Liyang Lu, Amr Hendy, Samyam Rajbhandari, Yuxiong He, and Hany~Hassan Awadalla.
\newblock Scalable and efficient moe training for multitask multilingual models.
\newblock \emph{arXiv preprint arXiv:2109.10465}, 2021.

\bibitem[Kudugunta et~al.(2021)Kudugunta, Huang, Bapna, Krikun, Lepikhin, Luong, and Firat]{Kudugunta2021BeyondDT}
Sneha Kudugunta, Yanping Huang, Ankur Bapna, Maxim Krikun, Dmitry Lepikhin, Minh-Thang Luong, and Orhan Firat.
\newblock Beyond distillation: Task-level mixture-of-experts for efficient inference.
\newblock In \emph{EMNLP}, 2021.

\bibitem[Lepikhin et~al.(2020)Lepikhin, Lee, Xu, Chen, Firat, Huang, Krikun, Shazeer, and Chen]{lepikhin2020gshard}
Dmitry Lepikhin, HyoukJoong Lee, Yuanzhong Xu, Dehao Chen, Orhan Firat, Yanping Huang, Maxim Krikun, Noam Shazeer, and Zhifeng Chen.
\newblock Gshard: Scaling giant models with conditional computation and automatic sharding.
\newblock \emph{arXiv preprint arXiv:2006.16668}, 2020.

\bibitem[Liu et~al.(2022)Liu, Kim, Muzio, and Hassan]{liu2022gating}
Rui Liu, Young~Jin Kim, Alexandre Muzio, and Hany Hassan.
\newblock Gating dropout: Communication-efficient regularization for sparsely activated transformers.
\newblock In \emph{International Conference on Machine Learning}, pp.\  13782--13792. PMLR, 2022.

\bibitem[Rajbhandari et~al.(2022)Rajbhandari, Li, Yao, Zhang, Aminabadi, Awan, Rasley, and He]{Rajbhandari2022DeepSpeedMoEAM}
Samyam Rajbhandari, Conglong Li, Zhewei Yao, Minjia Zhang, Reza~Yazdani Aminabadi, Ammar~Ahmad Awan, Jeff Rasley, and Yuxiong He.
\newblock Deepspeed-moe: Advancing mixture-of-experts inference and training to power next-generation ai scale.
\newblock In \emph{ICML}, 2022.

\bibitem[Vaswani et~al.(2017)Vaswani, Shazeer, Parmar, Uszkoreit, Jones, Gomez, Kaiser, and Polosukhin]{Vaswani2017AttentionIA}
Ashish Vaswani, Noam~M. Shazeer, Niki Parmar, Jakob Uszkoreit, Llion Jones, Aidan~N. Gomez, Lukasz Kaiser, and Illia Polosukhin.
\newblock Attention is all you need.
\newblock In \emph{NIPS}, 2017.

\bibitem[Wang et~al.(2020)Wang, Zhai, and Awadalla]{wang2020multi}
Yiren Wang, ChengXiang Zhai, and Hany~Hassan Awadalla.
\newblock Multi-task learning for multilingual neural machine translation.
\newblock \emph{arXiv preprint arXiv:2010.02523}, 2020.

\bibitem[Xiong et~al.(2020)Xiong, Yang, He, Zheng, Zheng, Xing, Zhang, Lan, Wang, and Liu]{Xiong2020OnLN}
Ruibin Xiong, Yunchang Yang, Di~He, Kai Zheng, Shuxin Zheng, Chen Xing, Huishuai Zhang, Yanyan Lan, Liwei Wang, and Tie-Yan Liu.
\newblock On layer normalization in the transformer architecture.
\newblock In \emph{ICML}, 2020.

\bibitem[Zoph et~al.(2022)Zoph, Bello, Kumar, Du, Huang, Dean, Shazeer, and Fedus]{zoph2022designing}
Barret Zoph, Irwan Bello, Sameer Kumar, Nan Du, Yanping Huang, Jeff Dean, Noam Shazeer, and William Fedus.
\newblock Designing effective sparse expert models.
\newblock \emph{arXiv preprint arXiv:2202.08906}, 2022.

\end{thebibliography}
\bibliographystyle{iclr2023_conference}

\clearpage

\appendix

\section{Quantization algorithms}


\subsection{Quantization techniques}
We try two quantization techniques, they are (i) linear quantization which is mapping quantized integer values and the original float value uniformly and (ii) log-based quantization from \citet{Aji2020CompressingNM} which maps integer and float ranges in a log scale. In both cases, we choose channel-wise quantization over matrix-wise quantization based on the experiment in Appendix \ref{app:chan-mat}.

\textbf{Linear quantization with absolute maximum.}
The first technique is linear quantization which, given a matrix $\mA$ and b bits, it encodes $\mA$ as follows: 
\begin{center} \label{eq1:linquant}
$\displaystyle \vs_{j} = \frac{2 \times \max(|\mA_{:,j}|)}{2^{b}-1}$ \\
$\displaystyle \mQ_{:,j}= \integer(\frac{\mA_{:,j}}{\vs_{j}})$ \\

\end{center}
where $s$ is the scaling factor which can be chosen per channel as shown or per the whole tensor. At inference time, the quantized $\mQ$ is dequantized back to $\mA^{'}$ with the scaling factor $\vs$ as follows:

\begin{center}
$\displaystyle \mA^{'}_{:,j}=\mQ_{:,j} \times \vs_{j}$ \\
\end{center}

\textbf{Log-scale quantization.}
The second technique is log-scale quantization where $1$ bit is kept for the sign and (${b-1}$) bits are used to encode the log-scaled values. Given a matrix $\mA$, the quantization formula is as follows:
\begin{center}
$\displaystyle \mP = sign({\mA})$ \\
$\displaystyle \mT =clip(\frac{|\mA|}{s}, 1, 2^{1-2^{b-1})}$ \\
$\displaystyle \mQ = \lceil log_2(\frac{2}{3}\mT) \rceil$  \\
\end{center}
where $s$ can be chosen in two ways, either (i) the absolute maximum or (ii) the optimal value to minimize the mean squared error (MSE) between the quantized and original values which is described in \citet{Aji2020CompressingNM}. We use the second algorithm which we observe a better accuracy with the quantization. At inference time, the quantized weight values are dequantized based on the formula as follows:
\begin{center}
$\displaystyle \mA^{'}= \mP\times s \times 2^\mQ$ \\
\end{center}

\begin{figure}[h]
    \centering
    \begin{subfigure}[b]{0.45\textwidth}
        \centering
        \includegraphics[width=1.0\textwidth]{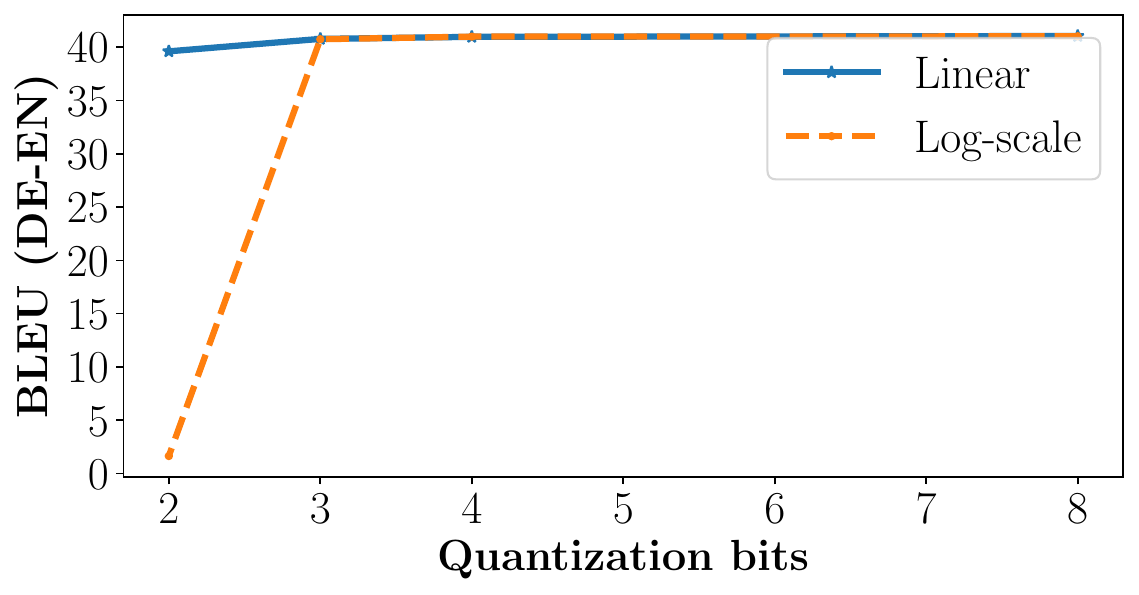}
        \vskip -0.1in
        \caption{\centering Expert FFNs}
        \label{fig:expert FFNs log_opt_s vs linear}
    \end{subfigure}
    \begin{subfigure}[b]{0.45\textwidth}
        \centering
        \includegraphics[width=1.0\textwidth]{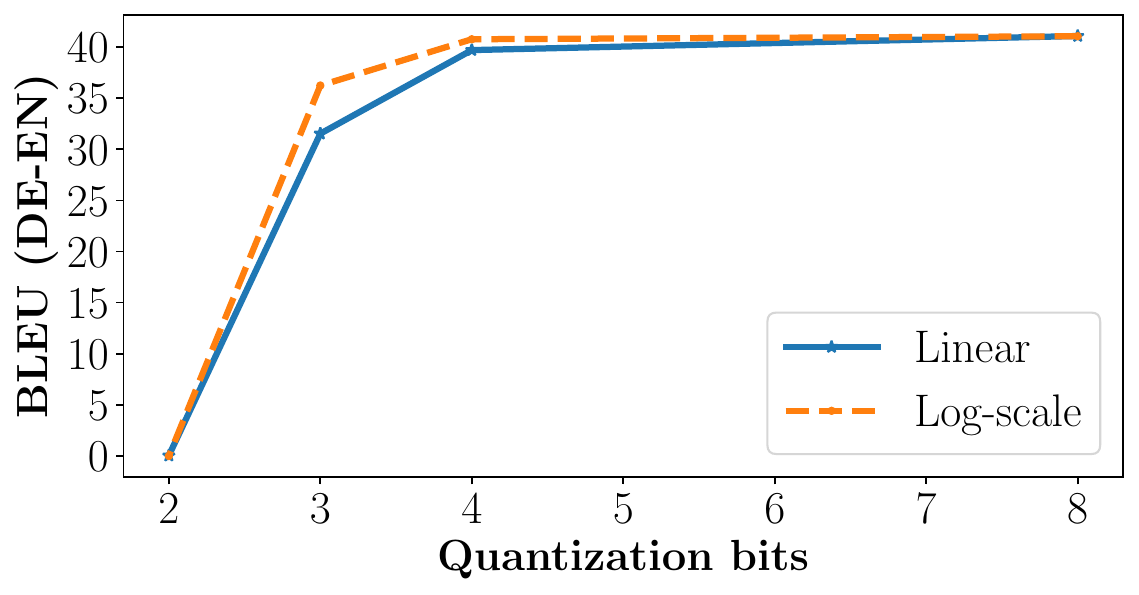}
        \vskip -0.1in
        \caption{\centering Dense FFNs}
        \label{fig:non expert FFNs log_opt_s vs linear}
    \end{subfigure}
    \caption{Linear quantization vs log-scale with optimal $\vs$ quantization.}
    \label{fig:linear-log}
\end{figure}

\section{Experimental setup}
\label{app:setup}
\textbf{Task.}
We use multilingual machine translation task for our experiments with two different dataset which are 20 language directions and 10 language directions respectively. We use sacrebleu~\footnote{https://github.com/mjpost/sacrebleu} on the detokenized output to measure the accuracy of the models. A single NVIDIA A100 running inside a docker container running Ubuntu 20.04 and CUDA 11.6 is used for all experiments, and all code is compiled with nvcc and gcc/g++ 9.3. We measure end-to-end runtime of the inference for the evaluation dataset.

\textbf{Datasets.}
We use two different datasets described below.
    For the larger dataset setting, we use internally collected dataset consists of 6 different languages which are German (de), French (fr), Italian (it), Spanish (es), Dutch (nl) and English (en). They are crawled from web, and each language pair has at least several hundred million sentences. We use 128,000 sub-words vocabulary built with sentencepiece\footnote{https://github.com/google/sentencepiece} library. The number of training sentences is included in Appendix \ref{app:datastat}. \\
    For the smaller dataset setting, we use WMT-10 benchmark dataset widely used for public benchmarks \citep{wang2020multi, kim2021scalable}. There are 32.5 million sentence pairs for English-centric 20 language pairs including French (fr), Czech(cs), German (de), Finnish (fi), Latvian (lt), Estonian (et), Romanian (ro), Hindi (hi), Turkish(tr) and Gujarati (gu).

\textbf{Model architecture.}
For all the experiments with large dataset, we use 24 transformer \citep{Vaswani2017AttentionIA} encoder layers and 12 transformer decoder layers following the deeper encoder and shallower decoder practice \citep{kim2019research, Kasai2021DeepES} to be more efficient at auto-regressive decoding. The embedding dimension is $1,024$ and FFN hidden dimension is $4,096$. For the positional information encoding to the hidden state, we use Transformer with Untied Positional Encoding (TUPE) proposed in \citet{Ke2021RethinkingPE} instead of the conventional sinusoidal positional embedding. Another design choice is the location of layer normalization. For the training stability, we use pre-layer normalization proposed in \cite{Xiong2020OnLN} instead of the original post-layer normalization from \citep{Vaswani2017AttentionIA}. We train MoE and dense models for the comparison. The model architecture choices mentioned here are common for both models. The only difference between dense and MoE models is the number of experts. We use 32 experts for the MoE model trained with the larger web data. We use beam search decoding with beam size of 5. For the experiments with smaller dataset, we use 12 transformer encoder layers and 6 transformer decoder layers. The embedding dimension is $768$ and FFN hidden dimension is $3,072$. In this setting, we use MoE layers with 128 experts at every other layer.

\textbf{MoE architecture.}
For the MoE model specific settings, we use top-1 learned gating from \cite{fedus2021switch} and use an MoE layer at every other layer which are even numbered layers \citep{lepikhin2020gshard, fedus2021switch, kim2021scalable}. During the training of MoE models, we use jittering noise and balancing loss (ratio of $0.01$) suggested in \cite{lepikhin2020gshard, fedus2021switch} to more uniformly distribute expert utilization. To prevent overfitting and better regularize the model, we use gating dropout ($0.2$) \citep{liu2022gating} as well.

\section{Channel-wise vs matrix-wise quantization}
\label{app:chan-mat}
\begin{figure}[h]
    \centering
    \includegraphics[width=0.7\textwidth]{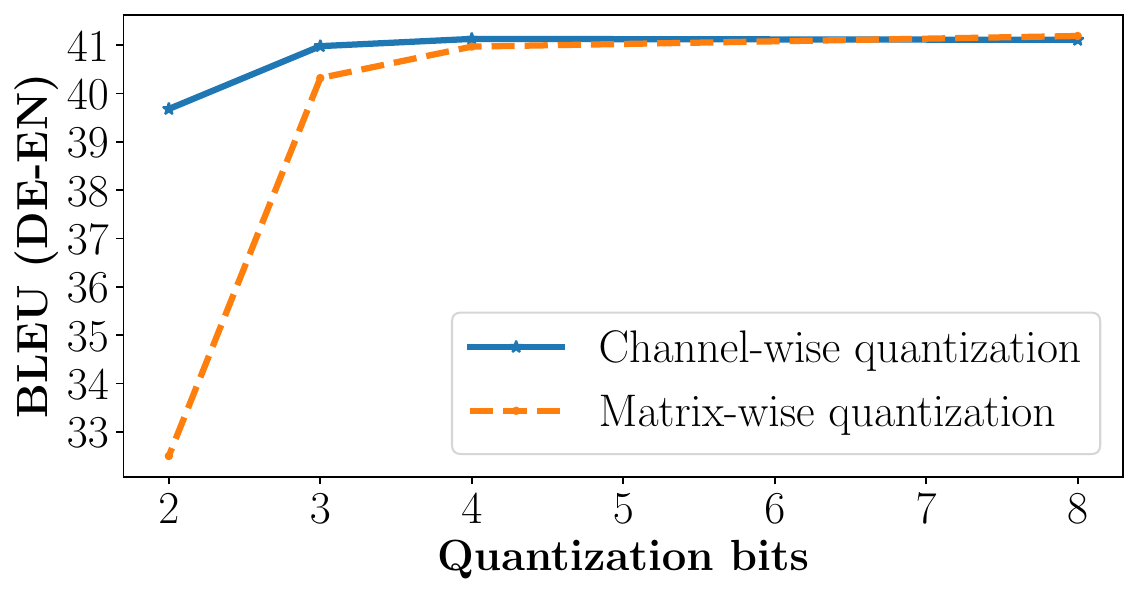}
    \vskip -0.1in
    \caption{Linear quantization of expert FFNs with channel-wise and matrix-wise scaling factors.}
    \label{fig:linear_channel_vs_tensor_expert_ffns}
\end{figure}

Scaling factors are calculated by the quantization algorithm and stored in half precision floating-point (fp16) numbers to dequantize the matrices with. These factors can be chosen on the channel scale or the whole matrix scale. As shown in figure \ref{fig:linear_channel_vs_tensor_expert_ffns}, channel-wise quantization gives quite higher scores than tensor-wise especially for low precision. Additional parameters to store channel-wise scaling factors is small, because only one value is needed for a channel and less than 1\% of total parameters in a matrix. Therefore, we use channel-wise quantization for all the quantization experiments.

\section{Quantization of different layers in a dense model}
\label{app:dense-layers}
In the paper, we compare a dense model and an MoE model in terms of quantization robustness. To make a fair comparison, we consider quantizing only half of the dense transformer blocks' FFNs, because we quantize expert weights only which exist only in every other block (even numbered). We compare three different configurations - (1) quantizing even numbered blocks' FFNs only, (2) quantizing odd numbered blocks' FFNs only and (3) quantizing all FFN layers. As can be seen in Figure \ref{app:dense-layers}, quantizing even numbered blocks' FFNs affects the accuracy the least, and quantizing all FFN layers give the worst result. Based on this experiment, we quantize only even numbered transformer blocks' FFNs for the dense model in all the experiments and comparisons.
\begin{figure}[h]
    \centering
    \includegraphics[width=0.7\textwidth]{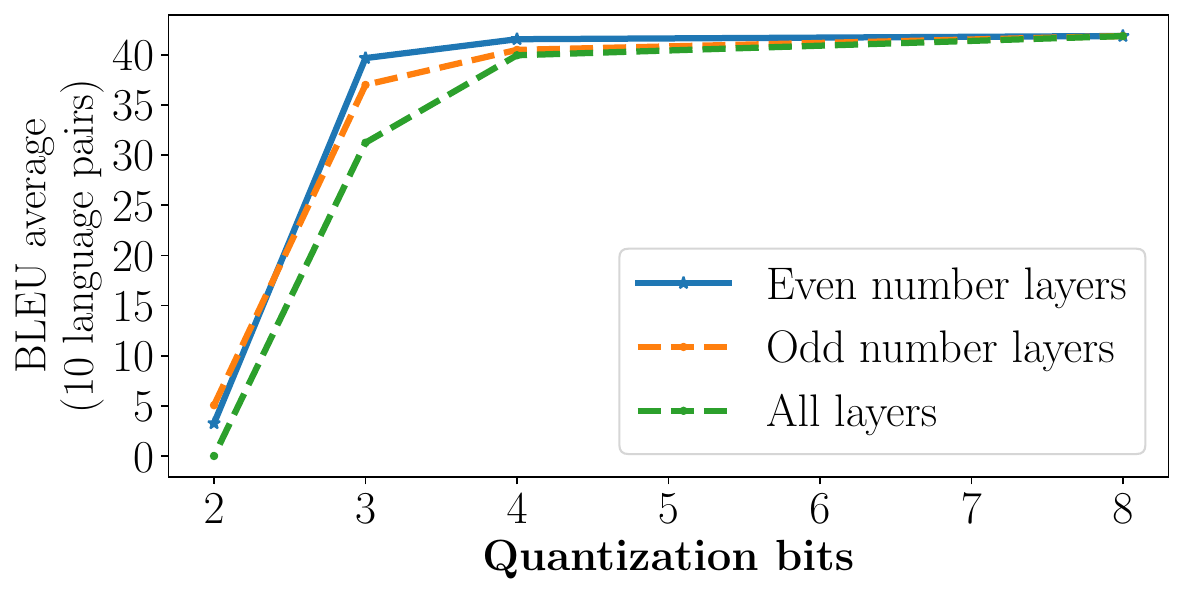}
    \vskip -0.1in
    \caption{Quantization impact of different layers in a dense model.}
    \label{fig:dense-layers}
\end{figure}

\section{Skewness of weight matrices in MoE and dense models}
\label{app:skew}
In the analysis of model weight distribution in Section \ref{sec:quant_methods}, we observe that dense models' FFN layers tend to have more outliers than MoEs' expert FFN layers. We measure the skewness of weight distribution of those in Table \ref{tab:skewness}.

\begin{table}[ht]
\caption{Expert vs non-expert FFN layers parameters distribution skewness}
\label{tab:skewness}
\begin{center}
\begin{small}
\begin{tabular}{cccccccccccc}
\hline
\multicolumn{1}{c}{\bf Parameter} & \multicolumn{1}{c}{\bf skew} \\

\hline
encoder expert 15 FFN fc1 layer 0 & -0.002 \\
\hline
encoder expert 15 FFN fc2 layer 0 &	-0.190 \\ 
\hline
encoder expert 15 FFN fc1 layer 6 & -0.002 \\
\hline
encoder expert 15 FFN fc2 layer 6 &	-0.002 \\
\hline
encoder non-expert FFN fc1 layer 1 & -0.019 \\ 
\hline
encoder non-expert FFN fc2 layer 1	& -10.729 \\ 
\hline
encoder non-expert FFN fc1 layer 7 & 0.003 \\ 
\hline
encoder non-expert FFN fc2 layer 7 & -1.574 \\
\hline
\hline

        encoder expert FFN fc1 mean & 0.00  \\ \hline
        encoder expert FFN fc2 mean & -0.63  \\ \hline
        decoder expert FFN fc1 mean & 0.00  \\ \hline
        decoder expert FFN fc2 mean & 0.48  \\ \hline
        encoder non-expert FFN fc1 mean & 0.00  \\ \hline
        encoder non-expert FFN fc2 mean & -1.84  \\ \hline
        decoder non-expert FFN fc1 mean & 0.00  \\ \hline
        decoder non-expert FFN fc2 mean & -0.09  \\ \hline
\end{tabular}
\end{small}
\end{center}
\end{table}

\section{Machine translation dataset summary}
\label{app:datastat}
Table \ref{tab:datastat} shows the number of parallel sentences used to train dense and MoE models. All languages have at least 300 million sentences and the differences in the number among languages are less than two times.

\begin{table}[ht]
\caption{The number of parallel sentences including backtranslation data.}
\label{tab:datastat}
\begin{center}
\begin{small}
\begin{tabular}{ccccccccc}
\hline
\multirow{2}{*}{\bf Language} & \multicolumn{2}{c}{\bf Number of parallel sentences (million)}        \\
 & \multicolumn{1}{c}{\bf xx $\rightarrow$ English} & \multicolumn{1}{c}{\bf English $\rightarrow$ xx}   \\
\hline
DE (German) & 505 & 411 \\
ES (Spanish) & 448 & 407  \\
FR (French) & 448 & 376  \\
IT (Italian) & 447 & 303  \\
NL (Dutch) & 302 & 378 \\
\hline
\end{tabular}
\end{small}
\end{center}
\end{table}



\section{Detailed BLEU score differences with quantization applied to the model trained on public WMT dataset}
\label{app:wmt}
Table \ref{tab:moe-dense-trained-on-wmt10} shows individual BLEU score changes with various quantization bits for MoE and dense models trained on public WMT dataset.

\begin{table}[t]
\caption{The BLEU score differences in percentage (\%) after quantization on different language pairs in WMT dataset. The rows with fp16 show the baseline BLEU scores.}
\label{tab:moe-dense-trained-on-wmt10}
\begin{center}
\begin{scriptsize}
\begin{tabular}{cccccccccccccc}
\hline
\multicolumn{1}{c}{\bf Bits}  & \multicolumn{1}{c}{\bf Model} & \multicolumn{1}{c}{\bf en-cs} & \multicolumn{1}{c}{\bf en-de} & \multicolumn{1}{c}{\bf en-et} & 
\multicolumn{1}{c}{\bf en-fi} & \multicolumn{1}{c}{\bf en-fr} & \multicolumn{1}{c}{\bf en-gu} & \multicolumn{1}{c}{\bf en-hi} & \multicolumn{1}{c}{\bf en-lv} & \multicolumn{1}{c}{\bf en-ro} & \multicolumn{1}{c}{\bf en-tr} & \multicolumn{1}{c}{\bf Avg.(en-xx)} \\ \hline
fp16 & Dense & 23.89 & 31.46 & 17.80 & 18.75 & 28.54 & 10.34 & 11.98 & 22.29 & 27.22 & 15.81 & 20.81 &\\ 
(BLEU) & MoE & 26.09 & 34.36 & 18.27 & 22.17 & 31.34 & 13.04 & 12.16 & 23.26 & 27.95 & 16.89 & 22.55
 \\ \hline

\multirow{2}{*}{8-bit} & Dense & -0.39 & -0.09 & -0.32 & 0.60 & 0.01 & -0.80 & 0.61 & -0.26 & 0.17 & -0.09 & -0.05  &\\
~ & MoE & 0.01   & -0.15   & 0.64   & -0.33   & 0.19   & 0.86   & 0.02   & -0.04   & -0.15   & -0.03  & 0.05 \\ \hline

\hline

\multirow{2}{*}{4-bit} & Dense  & -1.11 & -1.91 & -3.15 & -1.50 & 1.03 & -7.08 & -4.44 & -2.38 & -1.65 & -1.89  & -1.90
 \\ 
~ & MoE & -0.30 & -0.62 & 0.30 & -0.62 & -0.13 & -0.97 & 1.53 & -0.81 & -0.82 & -0.22  & -0.36 \\ \hline

\hline
\multirow{2}{*}{3-bit}  & Dense  & -10.87 & -7.86 & -12.87 & -11.70 & -3.96 & -32.03 & -24.76 & -11.16 & -7.05 & -12.74 & -11.24  & \\
~ & MoE & -0.84 & -1.06 & -1.79 & -1.97 & 0.35 & -2.80 & -0.70 & -1.98   & -1.05 & -1.64 & -1.21 \\ \hline
\multirow{2}{*}{2-bit} & Dense  & -97.44 & -86.29 & -91.79 & -91.02 & -85.75 & -98.26 & -96.48 & -94.14 & -87.30 & -95.02  & -91.21 & \\ 
& MoE & -8.84 & -9.15 & -17.06 & -13.24 & -5.62 & -25.24 & -16.38 & -16.11 & -11.04 & -14.48 & -12.34
\\ \hline
    \hline
\multicolumn{1}{c}{\bf Bits}  & \multicolumn{1}{c}{\bf Model} & \multicolumn{1}{c}{\bf cs-en} & \multicolumn{1}{c}{\bf de-en} & \multicolumn{1}{c}{\bf et-en} & \multicolumn{1}{c}{\bf fi-en} & \multicolumn{1}{c}{\bf fr-en} & \multicolumn{1}{c}{\bf gu-en} & \multicolumn{1}{c}{\bf hi-en} & \multicolumn{1}{c}{\bf lv-en} & \multicolumn{1}{c}{\bf ro-en} & \multicolumn{1}{c}{\bf tr-en} & \multicolumn{1}{c}{\bf Avg.(xx-en)}\\ \hline

fp16 & Dense & 29.48 & 35.62 & 23.43 & 23.91 & 31.89 & 16.54 & 14.97 & 26.25 & 35.68 & 18.52 & 25.63 & \\ 
(BLEU)& MoE & 31.25 & 38.21 & 23.67 & 25.64 & 32.59 & 19.55 & 15.89 & 25.22 & 34.80 & 20.27 & 26.71
\\ \hline

\multirow{2}{*}{8-bit} & Dense & 0.02   & -0.02   & 0.10   & -0.33   & -0.15   & -0.37   & -0.40   & 0.33   & -0.34   & 0.14   & -0.09  & \\ 
& MoE & 0.07   & 0.12   & 0.08   & 0.06   & -0.10   & 0.14   & -0.49   & -0.03   & 0.05   & -0.17   & 0.00   \\ \hline

\multirow{2}{*}{4-bit} & Dense & -0.24   & -0.78   & -3.74   & -1.72   & -1.69   & -4.58   & -0.56   & -1.97   & -0.15   & -1.84   & -1.53 &  \\ 
& MoE & 0.44   & 0.01   & -1.00   & 0.25   & -0.03   & 0.07   & 1.06   & -0.98   & 0.67   & -0.56   & 0.01   \\ \hline

\multirow{2}{*}{3-bit} & Dense & -7.25   & -7.11   & -10.44   & -10.36   & -6.44   & -18.67   & -16.68   & -11.52   & -7.39   & -10.39   & -9.68 &  \\ 
& MoE & -0.86   & -0.14   & -2.04   & -1.10   & 1.02   & -2.55   & 1.11   & -2.11   & -1.45   & -2.91   & -1.01   \\ \hline 

\multirow{2}{*}{2-bit} & Dense & -81.78   & -74.17   & -83.08   & -85.13   & -72.44   & -94.23   & -89.54   & -81.50   & -80.54   & -85.70   & -81.33  & \\ 
& MoE & -6.12   & -7.69   & -16.78   & -11.29   & -2.16   & -20.14   & -16.42   & -15.82   & -12.34   & -17.61 & -11.54 \\ \hline\hline
\end{tabular}
\end{scriptsize}
\end{center}
\end{table}

\section{Detailed BLEU score differences with quantization applied to 5.3B model.}
Table \ref{tab:moe-dense} shows individual BLEU score changes with various quantization bits for MoE and dense models measured with the internal validation dataset. Table \ref{tab:moe-dense-wmt10} shows the same model's evaluation performance on two WMT public dataset.

\begin{table}[t]
\caption{The BLEU score differences in percentage (\%) after quantization on different language pairs. The rows with fp16 show the baseline BLEU scores.}
\label{tab:moe-dense}
\begin{center}
\begin{small}
\begin{tabular}{cccccccccccc}
\hline
\multicolumn{1}{c}{\bf Quantization Bits} & \multicolumn{1}{c}{\bf Model} & \multicolumn{1}{c}{\bf de-en}&\multicolumn{1}{c}{\bf es-en}   & \multicolumn{1}{c}{\bf fr-en} & \multicolumn{1}{c}{\bf it-en}&\multicolumn{1}{c}{\bf nl-en} &\multicolumn{1}{c}{\bf Avg. (xx-English)}     \\
\hline
fp16 & Dense & 40.31 & 53.09 & 49.13 & 44.03 & 46.23 & 46.56 & \\ 
(Baseline BLEU)& MoE & 41.49 & 53.79 & 50.26 & 46.97 & 47.53 & 48.01\\ \hline
\hline
8-bit & Dense & -0.03 & -0.08 & -0.02 & 0.01 & -0.05 & -0.04\\
(\% difference)& MoE  & -0.10 & -0.06 & 0.00 & -0.02 & 0.03 & -0.03 \\
\hline
4-bit & Dense  & -0.78 & 0.29 & -0.23 & -0.93 & -0.20 &  -0.37\\
(\% difference)& MoE & -0.50 & -0.11 & -0.10 & -0.39 & -0.02 & -0.22\\
\hline
3-bit & Dense  & -6.36 & -2.51 & -4.24 & -5.93 & -2.67 & -4.34\\
(\% difference)& MoE & -0.92 & 0.26 & -0.26 & -1.26 & 0.29 & -0.38\\
\hline
2-bit & Dense  & -95.44 & -94.42 & -95.51 & -95.10 & -93.31 & -94.76 \\
(\% difference)& MoE  & -4.35 & -1.00 & -2.64 & -7.01 & -0.70 &  -3.14\\
\hline
\hline
 &  & \multicolumn{1}{c}{\bf en-de}& \multicolumn{1}{c}{\bf en-es}& \multicolumn{1}{c}{\bf en-fr} & \multicolumn{1}{c}{\bf en-it}&\multicolumn{1}{c}{\bf en-nl} &\multicolumn{1}{c}{\bf Avg. (English-xx)}    \\
\hline
fp16& Dense & 38.74 & 46.44 & 50.82 & 40.09 & 41.69 & 43.55 & \\
(Baseline BLEU)& MoE & 39.90 & 47.47 & 52.45  & 41.25 & 42.36 & 44.69 \\ \hline
\hline
8-bit & Dense   & -0.04 & -0.07 & 0.02 & -0.05 & 0.09 & -0.01 \\
(\% difference)& MoE   & 0.05 & -0.01 & -0.03 & 0.00 & 0.00 & 0.00 \\
\hline
4-bit & Dense  & -0.76 & -1.11 & -0.29 & -0.70 & -0.26 & -0.62\\
(\% difference)& MoE   & 0.31 & -0.90 & -0.74 & -0.45 & -0.68 & -0.49\\
\hline
3-bit & Dense  & -5.82 & -4.79 & -3.96 & -5.41 & -4.54 & -4.91 \\
(\% difference)& MoE & -0.21 & -2.12 & -1.41 & -0.87 & -0.89 & -1.10 \\
\hline
2-bit & Dense   & -97.28 & -96.16 & -95.52 & -96.68 & -94.83 & -96.09 \\
(\% difference)& MoE  & -5.24 & -6.19 & -5.19 & -5.30 & -4.48 & -5.28 \\
\hline
\end{tabular}
\end{small}
\end{center}
\end{table}


\begin{table}[t]
\caption{The BLEU score differences in percentage (\%)  of 5.3B MoE model after quantization on different language pairs on WMT datasets. The rows with fp16 show the baseline BLEU scores.}
\label{tab:moe-dense-wmt10}
\begin{center}
\begin{small}
\begin{tabular}{ccccccc}
\hline
\multicolumn{1}{c}{\bf Quantization Bits} & \multicolumn{1}{c}{\bf Model} & \multicolumn{1}{c}{\bf de-en}&\multicolumn{1}{c}{\bf fr-en} &\multicolumn{1}{c}{\bf Avg. (xx-English)}     \\
\hline
fp16 & Dense & 50.11 & 42.98 & 46.54 & \\
(Baseline BLEU) & MoE & 52.73 & 44.04 & 48.39 & \\ \hline
8-bit & Dense & 0.04 & 0.11  & 0.07   & \\
(\% difference)& MoE & 0.09 & -0.04 & 0.03 & \\
\hline
4-bit & Dense & -0.59 & -1.27  & -0.91 &\\
(\% difference)& MoE & -0.47 & -0.36 & -0.42 & \\
\hline
3-bit & Dense  & -5.75 & -6.17  & -5.94 & \\
(\% difference)& MoE & -1.15 & -0.90 & -1.03 &\\
\hline
2-bit & Dense  & -96.88 & -95.59  & -96.28 & \\
(\% difference)& MoE & -5.37 & -3.68 & -4.60 &\\
\hline
\hline
\multicolumn{1}{c}{\bf} & \multicolumn{1}{c}{\bf } & \multicolumn{1}{c}{\bf en-de}&\multicolumn{1}{c}{\bf en-fr} &\multicolumn{1}{c}{\bf Avg. (English-xx)}     \\
\hline
fp16 & Dense & 50.90 &	44.47 & 47.68 & \\
(Baseline BLEU)& MoE & 52.90 & 45.51 & 49.21 & \\ \hline
8-bit & Dense & 0.00 & 0.02  & 0.01 & \\
(\% difference)& MoE & -0.05 & 0.23 & 0.08 & \\
\hline
4-bit & Dense & 0.24 & -1.31 & -0.48
& \\
(\% difference)& MoE  & -0.93 & 0.25 & -0.39 &\\
\hline
3-bit & Dense & -5.86 & -7.53  & -6.64 & \\
(\% difference)& MoE  &  -1.41 & -0.69 & -1.08 &\\
\hline
2-bit & Dense & -97.77 & -96.22  & -97.05 & \\
(\% difference)& MoE  &  -6.34 & -6.15 & -6.25 &\\
\hline
\end{tabular}
\end{small}
\end{center}
\end{table}


\end{document}